\documentclass[final]{cvpr}

\usepackage{times}
\usepackage{epsfig}
\usepackage{graphicx}
\usepackage{amsmath}
\usepackage{amssymb}

\usepackage{booktabs}
\usepackage{tabularx}

\usepackage{graphicx}

\usepackage{float}
\usepackage{multirow}
\usepackage{diagbox}

\usepackage{cuted}
\usepackage{capt-of}
\usepackage{pifont}
\usepackage{color}

\usepackage{makecell}

\usepackage[pagebackref=true,breaklinks=true,colorlinks,bookmarks=false]{hyperref}

\def\cvprPaperID{4703} 
\def\confYear{CVPR 2021}

\makeatletter
\newcommand{\printfnsymbol}[1]{%
  \textsuperscript{\@fnsymbol{#1}}%
}
\makeatother

\begin{document}

\title{Progressively Complementary Network for Fisheye Image Rectification Using Appearance Flow}

\author{Shangrong Yang\thanks{Equal Contributions;} , Chunyu Lin\printfnsymbol{1}\thanks{Corresponding author: cylin@bjtu.edu.cn} , Kang Liao\printfnsymbol{1}, Chunjie Zhang, Yao Zhao\\
Institute of Information Science, Beijing Jiaotong University\\
Beijing Key Laboratory of Advanced Information Science and Network, Beijing, 100044, China\\
{\tt\small \{sr\_yang, cylin, kang\_liao, cjzhang, yzhao\}@bjtu.edu.cn}
}

\maketitle
\pagestyle{empty}
\thispagestyle{empty}

\begin{abstract}
   Distortion rectification is often required for fisheye images. The generation-based method is one mainstream solution due to its label-free property, but its naive skip-connection and overburdened decoder will cause blur and incomplete correction. First, the skip-connection directly transfers the image features, which may introduce distortion and cause incomplete correction. Second, the decoder is overburdened during simultaneously reconstructing the content and structure of the image, resulting in vague performance. To solve these two problems, in this paper, we focus on the interpretable correction mechanism of the distortion rectification network and propose a feature-level correction scheme. We embed a correction layer in skip-connection and leverage the appearance flows in different layers to pre-correct the image features. Consequently, the decoder can easily reconstruct a plausible result with the remaining distortion-less information. In addition, we propose a parallel complementary structure. It effectively reduces the burden of the decoder by separating content reconstruction and structure correction. Subjective and objective experiment results on different datasets demonstrate the superiority of our method.
\end{abstract}
\vspace{-0.3cm}
\section{Introduction}

Currently, fisheye cameras are widely used in video surveillance \cite{Muhammad2019EfficientDC}, autonomous driving \cite{Geiger2012AreWR} and mobile applications \cite{Mahmoodi2019OptimalJS}. However, the images captured by fisheye cameras are not suitable for most computer vision techniques designed for perspective images, such as target tracking \cite{tracking}\cite{Li2019TargetAwareDT}, motion estimation \cite{motionestimate}\cite{Bao2019MEMCNetME}, scene segmentation \cite{Scene_Seg}\cite{Fu2019DualAN}. In order to resolve the contradiction, distortion rectification has drawn great attention for decades.

\begin{figure}[!t]
\centering
\includegraphics[scale=0.13]{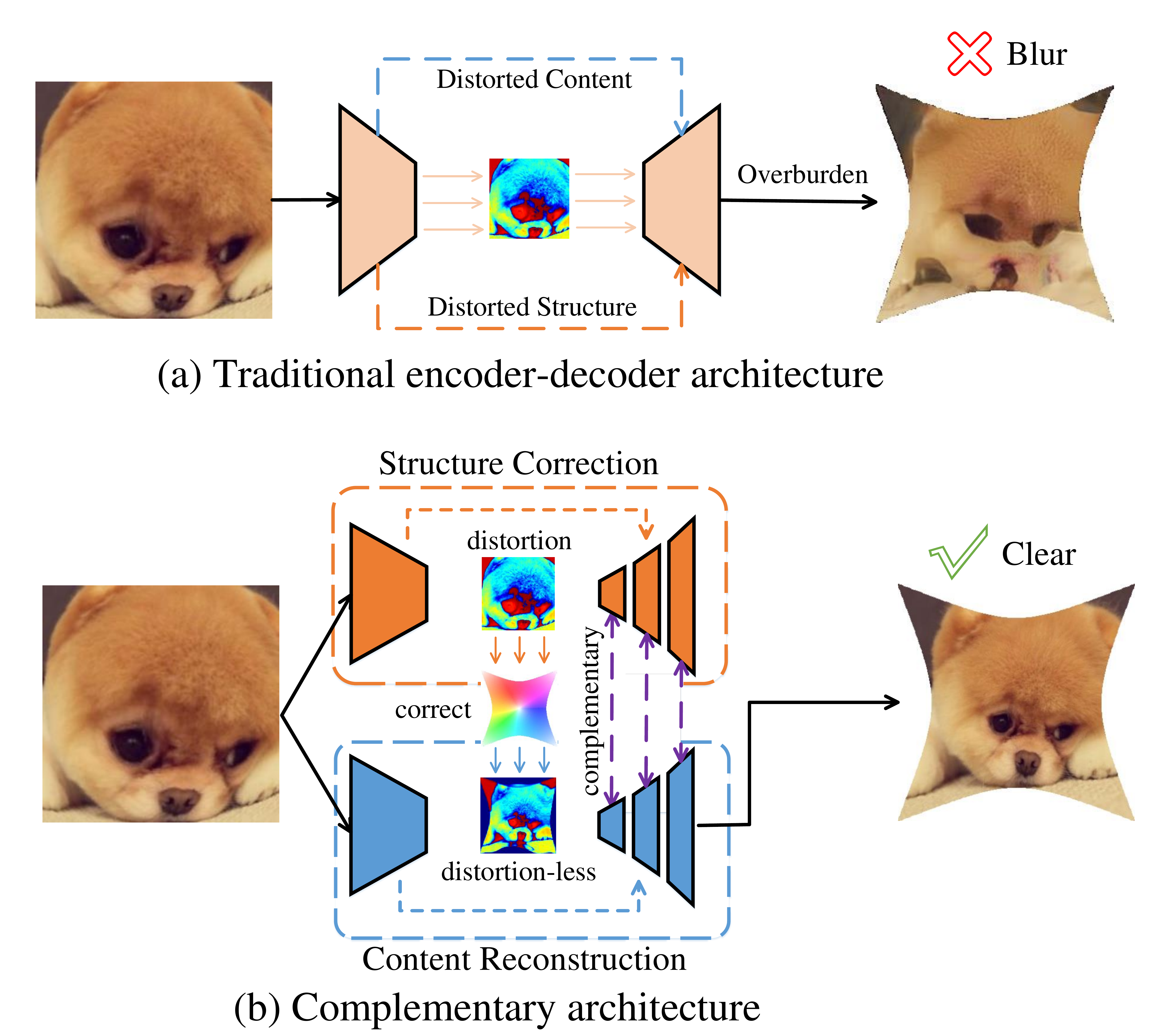}
\caption{\textbf{Generation-based approaches for image rectification.} (a) Distorted features are utilized for image reconstruction directly. (b) Distorted features have a pre-correction by a predicted appearance flow before reconstructing the corrected image.}
\label{Blue_Unet}
\vspace{-0.7cm}
\end{figure}

Traditional algorithms \cite{Rui2014Unsupervised}\cite{Bukhari2013Automatic}\cite{Zhang2015}\cite{Barreto2005} automatically extract pervasive features to calculate corresponding model parameters. However, the number of detected features is unstable, which greatly influences the model accuracy. To solve this problem, many existing approaches \cite{Rong2016Radial}\cite{DeepCalib}\cite{Yin2018FishEyeRecNet}\cite{Xue2019}\cite{Liao2019}\cite{DDM} leverage the potential of deep learning which can roughly be divided into two categories: regression-based method and generation-based method. The regression-based methods \cite{Rong2016Radial}\cite{DeepCalib}\cite{Yin2018FishEyeRecNet}\cite{Xue2019} utilize convolutional neural network (CNN) to predict complex nonlinear model parameters. However, they have to trade-off the number of parameters in the nonlinear model. In contrast, the generation-based methods \cite{Liao2019}\cite{DDM}\cite{Blind} directly generate corrected images with the help of encoder-decoder structure. Nevertheless, the effects inferred by this structure have never been explored. As we can see from Fig. \ref{Blue_Unet}\textcolor{red}{a}, the skip-connection in structure communicates redundant information, such as distorted features extracted by the encoder, thus confounding the decoder. The transmitted distorted features present difficulties for image reconstruction, which is termed as the distortion diffusion problem.

\begin{figure}[!t]
\centering
\includegraphics[scale=0.24]{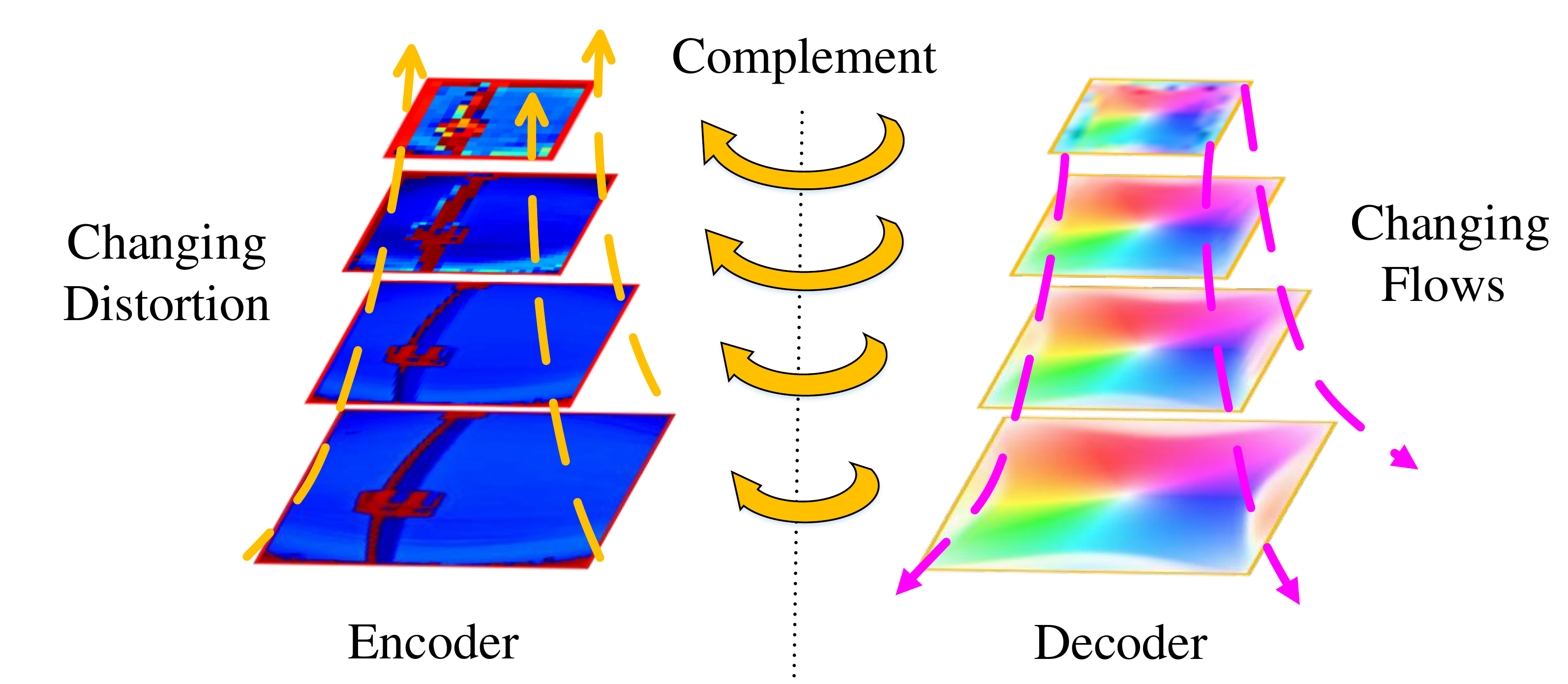}
\caption{\textbf{Gradual generation characteristic.} From low-level to high-level, the distortion of encoder features shows a gradual slight reduction, as well as the displacement of decoder flows.}
\label{complement}
\vspace{-0.5cm}
\end{figure}

In this paper, we propose a feature-level distortion rectification network\footnote{Available at \url{https://github.com/uof1745-cmd/PCN}}, which separates the structure correction and content reconstruction, as shown in Fig. \ref{Blue_Unet}\textcolor{red}{b}. It contains two modules: flow estimation and distortion correction. First, the flow estimation module estimates the distorted image structure and represents the result with dense appearance flows. Second, the distortion correction module leverages flows to correct the distorted features and uses the corrected features to reconstruct a plausible result. To bridge the two modules, we introduce a progressively complementary mechanism to achieve multi-level correction, as shown in Fig. \ref{complement}\textcolor{red}. According to our observation, from low-level to high-level, the encoder features on the distortion correction module show a gradual slight reduction in distortion. In the meantime, the decoder outputs at the flow estimation module also decrease progressively on displacement, which is termed as gradual generation characteristic. Therefore, the flow of each layer at the decoder can be used to correct corresponding feature maps, thus solving the problem of distortion diffusion and enhancing the performance. In addition, we propose a multi-scale loss for better supervision of corrected features. Experimental results show that our proposed method obtains superior performance, compared with state-of-the-art methods.

We summarize our contribution as follows:
\begin{itemize}
\item Feature-level distortion rectification scheme is proposed for the first time. Feature correction layers are embedded in skip-connection for feature pre-correction, which helps the decoder to reconstruct a plausible result.

\item The proposed unsupervised flow estimation module is able to estimate the distorted image structure. It can be trained in an end-to-end manner like a self-attention module \cite{STM}.

\item Taking advantage of the gradual generation characteristic, the correction in our network is progressive and complementary. Moreover, a multi-scale loss is introduced to supervise the corrected features.

\end{itemize}

\section{Related Work}
Distortion rectification plays an effective role to bridge the fisheye images and computer vision technology. Traditional methods \cite{ZHANG1999}\cite{Mei2007}\cite{Gasparini2009}\cite{Puig2010}\cite{Zhang2000AFN} can complete calibration by finding the corresponding feature points from different perspectives. However, such methods required special chessboards and human intervention. Therefore, automatic correction methods \cite{Dansereau2013DecodingCA}\cite{Stuiver1993Extended1D}\cite{Zhang2015}\cite{Barreto2005}\cite{Chander2009SummaryOC}\cite{Geiger2012AutomaticCA} had been made to solve these problems. Depending on the principle that straight lines have to be straight \cite{StraightLine}, they leveraged special detection methods to detect characteristic curves and then obtained the distortion parameters by calculating the curvature of curves. However, it was vulnerable due to the unstable number of characteristics. Deep learning methods \cite{Guo2017OnCO}\cite{Zhang2015LargeScaleAS}\cite{Lee2019CorrectionOB}\cite{Yang2010CorrectAP}\cite{Grant2015AutomaticEA}\cite{Ophus2016CorrectingND} solved the severe problems that remain in traditional automatic correction methods. Particularly, according to different networks, we categorized deep learning methods into two types, regression-based methods and generation-based methods.

\noindent\textbf{Regression-Based Methods.} Regression-based methods utilized a convolutional neural network(CNN) \cite{Krizhevsky2017ImageNetCW} to predict complex nonlinear model parameters. Rong et al. \cite{Rong2016Radial} pioneered to train the network on fitted data and used AlexNet to correct the distorted images. However, the limited discrete interval of parameters caused the trained network to perform poorly on complex fisheye images. Yin et al. \cite{Yin2018FishEyeRecNet} proposed a multicontext collaborative network, but semantic features can only provide limited guidance because of high dimensional features. Xue et al. \cite{Xue2019} imposed explicit geometry constraints to improve the network perception of distorted images. Although achieving better performance, it required a vast amount of labels, such as edge labels, distortion parameter labels, and normal images. Besides, the edge estimation network needed to be pre-trained, which brings a more complex operation.

\noindent\textbf{Generation-Based Methods.} The corrected image was directly generated with the help of a generative adversarial network(GAN) \cite{Ledig2017PhotoRealisticSI}. DR-GAN \cite{Liao2019} was the first adversarial framework for radial distortion rectification. It can directly learn the distribution pattern between distorted images and normal images instead of estimating the parameters. It achieved label-free training and one-stage rectification. However, the network was overburdened for rebuilding image content and structure simultaneously. The image content was blurred and the structure cannot be completely corrected. Liao et al. \cite{DDM} proposed a model-free distortion rectification framework for the single-shot case, bridged by the distortion distribution map. It yielded a more accurate correction on the distorted structure. However, cascade network \cite{DDM} caused image details lost easily, and general skip-connection led to distortion diffusion.
\begin{figure*}[!t]
\centering
\includegraphics[scale=.160]{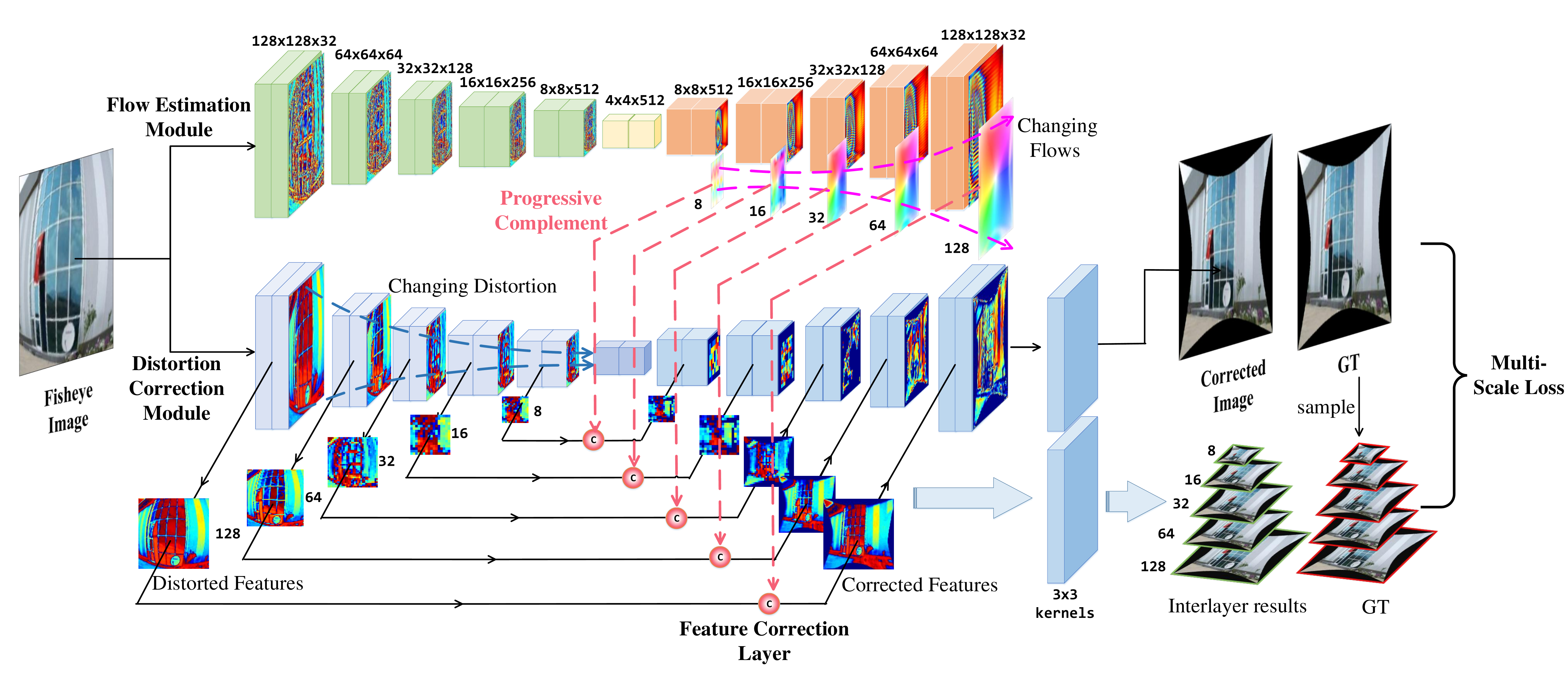}
\caption{
\label{structure}
Overview of our complementary network. The network architecture is composed of a flow estimation module (top) and a distortion correction module (bottom). The flow estimation module estimates the structure of the distorted image and provides a series of flows on each decoder layer. The distortion correction module leverages flows to correct corresponding distorted features in the correction layer. The corrected image features are supervised by a multi-scale loss to enhance the performance.}
\vspace{-0.3cm}
\end{figure*}

\section{Fisheye Models for Synthetic Data}
Obtaining massive real distorted images and their corresponding labels are labor-intensive. Therefore, generating synthetic distorted images for training with the fisheye camera model \cite{Rong2016Radial}\cite{Yin2018FishEyeRecNet}\cite{Xue2019}\cite{DDM} has become a mainstream approach. Generally, division model \cite{Fitzgibbon2001} and polynomial model \cite{Basu1995} are the most popular types. In image coordinate system, the Euclidean distance between an arbitrary point $P_{u}\left(x, y\right)$ and image center $P_{0}\left(x_{0}, y_{0}\right)$ on perspective image can be represented as $r_{u}$. $P_{u}$ has a corresponding point $P_{d}\left(x_{d}, y_{d}\right)$ in fisheye image. Similarly, the Euclidean distance between $P_{d}$ and distortion center is labeled as $r_{d}$. The mapping relationship between $r_{u}$ and $r_{d}$ can be represented by division model \cite{Fitzgibbon2001} as follows
\begin{equation}
r_{u}=\frac{r_{d}}{1+\sum_{i=1}^{n} k_{i} r_{d}^{2 i-1}}
\end{equation}
Where $k_{i}$ is distortion parameter. The distortion degree of fisheye image can be variable by changing the value of $k_{i}$. $n$ is the number of parameters. Generally, the larger the $n$ is, the more complex distortion state could be represented by polynomial. Compared with division model, polynomial model \cite{Basu1995} is more special by involving the angle of incident light. The polynomial model is usually expressed as follows
\begin{equation}
\theta_{u}=\sum_{i=1}^{n} k_{i} \theta_{d}^{2 i-1}, \qquad n=1,2,3,4, \ldots
\end{equation}
$\theta_{u}$ represents the angle of incident light and $\theta_{d}$ is the angle that light pass through the lens. Generally, $r_{d}$ and $\theta_{d}$ satisfy the equidistant projection relation, in which $r_{d}=f \theta_{d}$. $f$ is the focal length of fisheye camera. As for the pinhole camera, the projection model corresponds to $r_{u}=f \tan \theta_{u}$. We simplify the formula and get $\theta_{u}=\arctan \left(\frac{r_{u}}{f}\right) \approx \frac{r_{u}}{f}$. Therefore, we can calculate the relationship between $r_{u}$ and $r_{d}$ on polynomial model
\begin{equation}
r_{u}=f \sum_{i=1}^{n} k_{i} r_{d}^{2 i-1}, \qquad n=1,2,3,4, \ldots
\end{equation}

We merge the $k_{i}$ and $f$ to get the final polynomial model
\begin{equation}
r_{u}=\sum_{i=1}^{n} k_{i} r_{d}^{2 i-1}, \qquad n=1,2,3,4, \ldots
\end{equation}

In this paper, the polynomial model is selected to generate the synthesized fisheye images.
\section{Proposed Approach}
\subsection{Network Architecture}
Existing generation-based methods simultaneously reconstruct the structure and content of the image on the decoder, which leads to overburden for the decoder and causes blurred results. We separate structure correction and content reconstruction into two modules. As shown in Fig. \ref{structure}, our network consists of two main components, appearance flow estimation module and distortion correction module. Given a fisheye image with a size of $256\times256$, we fed it into two modules simultaneously. The appearance flow estimation module evaluates the distortion degree and presents it as appearance flows. The distortion correction module extracts features on the encoder. Since encoder features contain distortion, we treat them as distorted features. Each layer features are sent to the feature correction layer, leveraging the corresponding appearance flows to pre-correct. Thereafter, with the help of the corrected features, the decoder can concentrate on content reconstruction.

\noindent\textbf{Appearance Flow Estimation Module.} Benefiting from the encoder-decoder structure, this module utilizes it to extract features and generate a series of appearance flows. Specially, the output of each decoder layer is involved according to the progressively complementary mechanism that will be detailed later. The output features are experienced additional convolution with $3\times3$ kernels to obtain two-channel appearance flows. In this way, we obtain 5 appearance flows with sizes of 128, 64, 32, 16, 8. The process can be expressed as
\begin{equation}
\left \{ A_{f}^{i} \right \}_{i=1}^{5}=G_{s}\left ( I_{in} \right )
\end{equation}
Where $I_{in}$ is the input fisheye image. $G_{s}$ presents the appearance flow estimation module. $A_{f}^{i}$ is the output appearance flow of the i-th decoder layer.

\noindent\textbf{Feature Correction Layer.} To solve the distortion diffusion problem, we insert feature correction layer in skip connection, intending to pre-correct the image features before delivering. The predicted flow $A_{f}^{i}$ is leveraged to perform spatial transformation \cite{STM} as follow
\begin{equation}
I_{c}^{i}\left ( u,v \right )=I_{f}^{i}\left ( u+A_{f}^{i}\left ( u \right ),v+A_{f}^{i}\left ( v \right ) \right )
\end{equation}
$I_{f}^{i}$ is the i-th layer feature map in distortion correction module, and the corrected feature map is $I_{c}^{i}$.

\begin{figure}[!t]
\centering
\includegraphics[scale=.21]{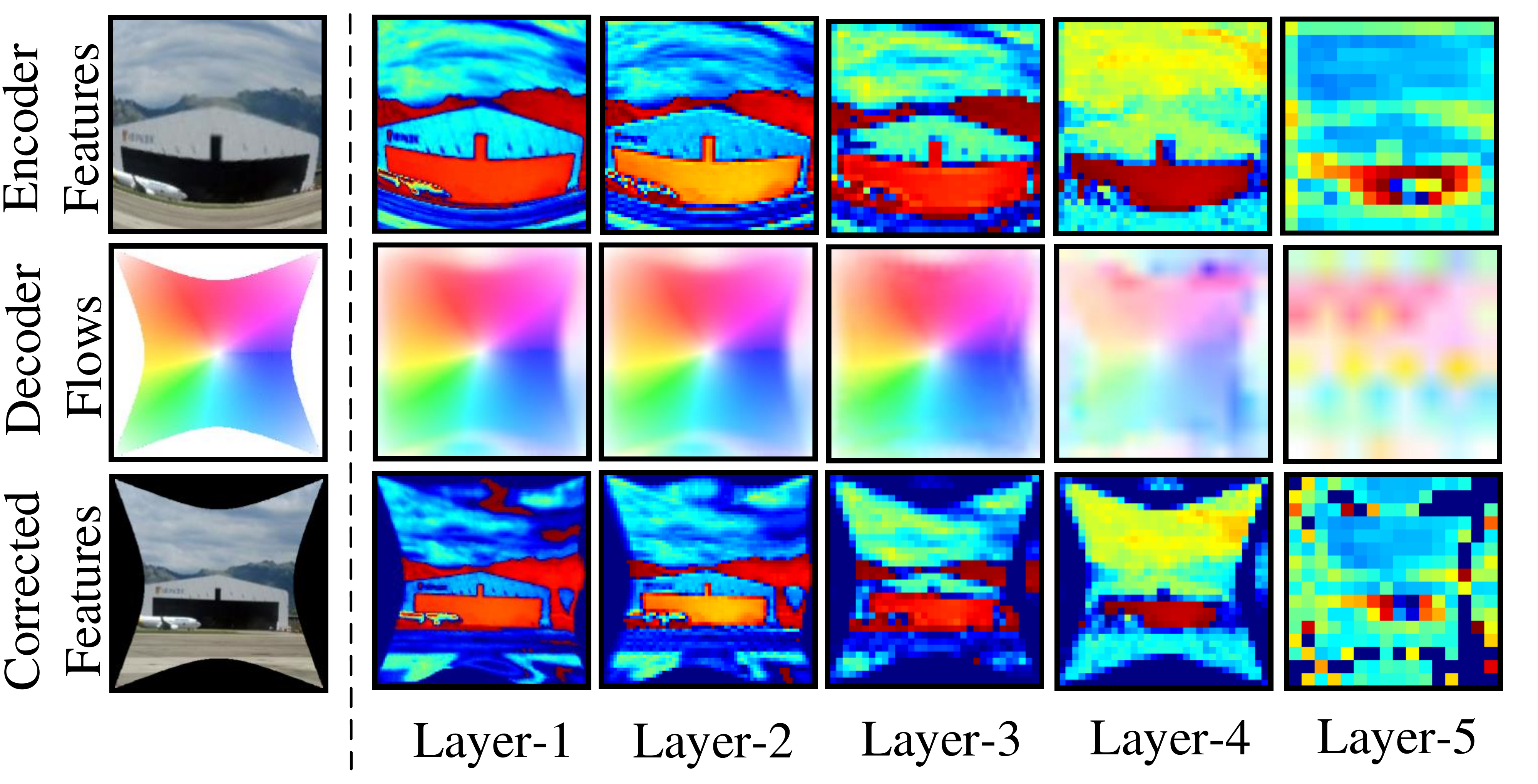}
\caption{
\label{Progress}
\textbf{Progressively complementary mechanism.} The progressive changing flows are leveraged to correct the progressive changing distortion features.}
\vspace{-0.3cm}
\end{figure}

\noindent\textbf{Progressively Complementary Mechanism.} Based on the visualizing of the feature maps, we impose a progressively complementary mechanism. As shown in Fig. \ref{Progress}, at the encoder of distortion correction module, continuous convolution and pooling operations blur the feature maps edges, making the degree of distortion appear to be slightly reduced. At the decoder of the flow estimation module, the distorted features transferred by skip-connection forces the predicted flows to have a greater displacement to represent greater distortion. Intuitively, we can represent the degree of distortion and displacement as follows
\begin{equation}
\begin{aligned}
c
& \geq D\left( I_{f}^{1} \right )\geq D\left ( I_{f}^{2} \right )\geq D\left ( I_{f}^{3} \right ) \\
& \geq D\left( I_{f}^{4} \right )\geq D\left ( I_{f}^{5} \right )\geq 0
\end{aligned}
\end{equation}
\begin{equation}
\begin{aligned}
k\cdot c
& \geq M\left( A_{f}^{1} \right )\geq M\left ( A_{f}^{2} \right )\geq M\left ( A_{f}^{3} \right ) \\
& \geq M\left( A_{f}^{4} \right )\geq M\left ( A_{f}^{5} \right )\geq 0
\end{aligned}
\end{equation}
Where $D$ is a function of that estimate the degree of input feature distortion. $M$ is a function of estimating the displacement degree of appearance flow. The bigger the distortion of an image, the greater the displacement is required for correction. Moreover, $c$ and $k$ are constant and we have $M\left ( A_{f}^{i} \right )_{max}/D\left ( I_{f}^{i} \right )_{max}=k$.

As seen from the formula, there is a significant hierarchical correspondence between $I_{f}^{i}$ and $A_{f}^{i}$. They share the same size with 128, 64, 32, 16, 8. Besides, powerful learning ability of the network can build a relationship with $D\left ( I_{f}^{i} \right ) \propto M\left ( A_{f}^{i} \right )$, which is progressive complement. Thereafter, we can leverages $A_{f}^{i}$ in feature correction layer to correct $I_{f}^{i}$.

\begin{figure}[!t]
\centering
\includegraphics[scale=.18]{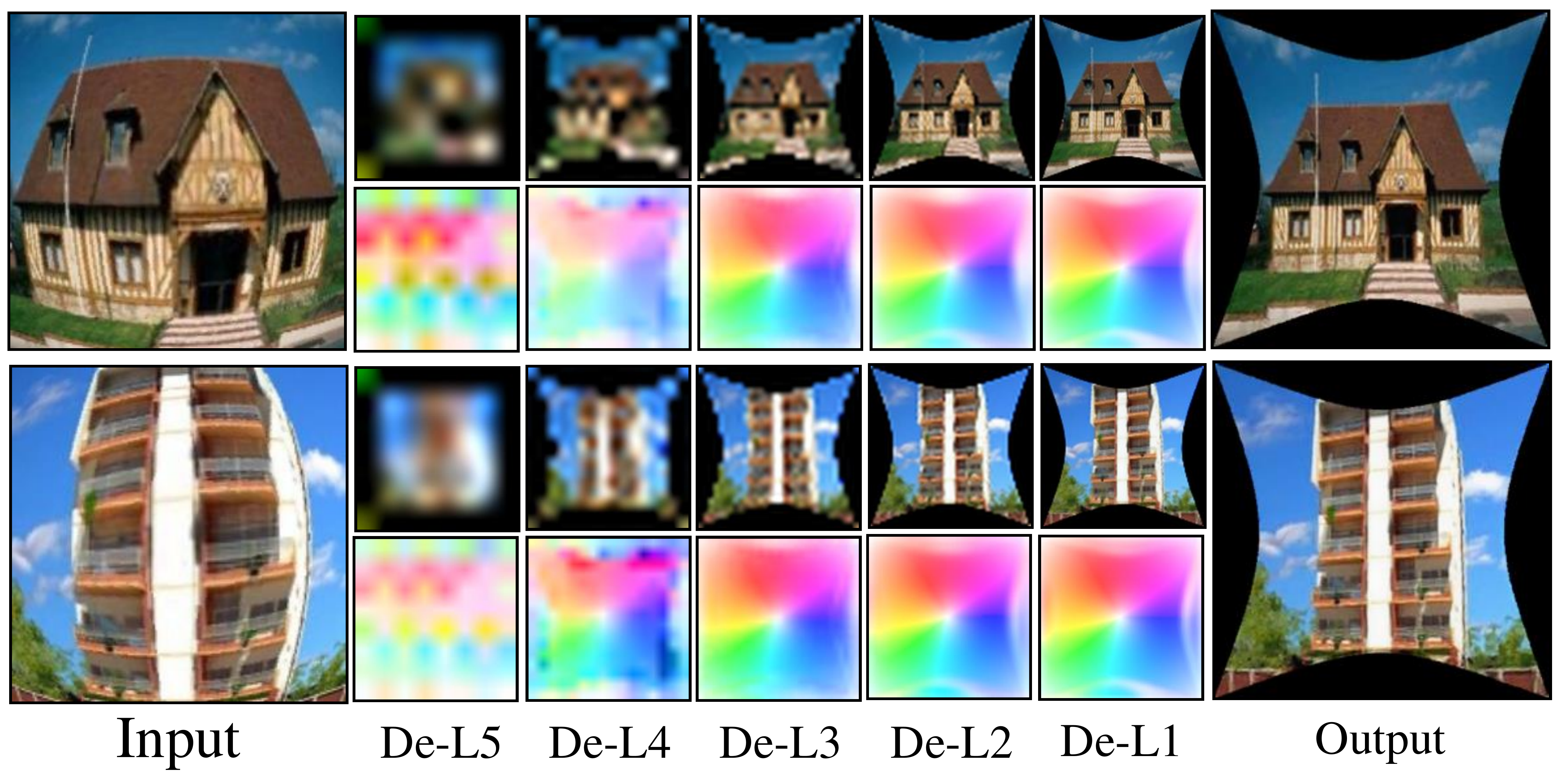}
\caption{
\textbf{Multi-scale corrected images.} With the help of progressively complementary mechanism, the feature map on each decoder layer is pre-corrected.}
\label{Multi-loss}
\vspace{-0.5cm}
\end{figure}

\noindent\textbf{Distortion Correction Module.} With the help of the progressively complementary mechanism, the feature maps used for concatenation on the decoder have been roughly corrected. The corrected feature map can bring a lot of details, without transmitting the distortion structure. Therefore, the network can generate a more visually realistic corrected image. The process of this module can be denoted as
\begin{equation}
\left \{ I_{out}^{i} \right \}_{k=1}^{6}=G_{c}\left ( I_{in},\left \{ A_{f}^{i} \right \}_{k=1}^{5} \right )
\end{equation}
Where $I_{out}^{i}$ denotes the i-th layer corrected image. $G_{c}$ indicates the distortion correction module. To ensure the correction quality, we send the concatenated features to the convolutional layer with $3\times3$ kernels to obtain multi-scale corrected images and downsample the ground truth image at the same scales to supervise them. The multi-scale corrected images can be shown in Fig. \ref{Multi-loss}. As a result, the integrated network can achieve better end-to-end training.

\subsection{Training strategy}
\label{strategy}
Our progressively complementary network is obtained by paralleling two subnetworks. Considering the complexity in the complementary network, we propose a progressively complementary training strategy that contains multiple loss functions to achieve stable end-to-end training.

\noindent\textbf{Reconstruction Loss.} Generally, the corrected image not only needs to have a normal structure but also needs to have better image details. Therefore, we first utilize the reconstruction loss to ensure the structural similarity between the corrected image and ground true image. We denote the corrected image as $I_{out}$. The ground true corrected image is $I_{gt}$. Therefore, reconstruction loss can be formulated as
\begin{equation}
\mathcal{L}_{r}=\left \| I_{out} - I_{gt}\right \|_{1}
\end{equation}

\noindent\textbf{Adversarial Loss.} Reconstruction Loss greatly helps generate the structure of the image, but it is powerless in generating the texture details. Subsequently, we use adversarial loss to enhance the image texture. Adversarial loss can be represented as follow
\begin{equation}
\begin{aligned}
\mathcal{L}_{adv}=\underset{G_{c}}{min}\underset{D}{max} & \left ( {E\left [ logD\left ( I_{gt} \right ) \right ] + } \right. \\
& \left. {E\left [ log\left ( 1-D\left ( G_{c}\left ( I_{in} \right ) \right ) \right ) \right ]}\right )
\end{aligned}
\end{equation}
Where $G_{c}$ and $D$ denotes the generator and discriminator in distortion correction module respectively.

\noindent\textbf{Enhanced Loss.} We introduce an enhanced loss to further enrich texture details. Specifically, enhanced loss consists of content loss and style loss\cite{StyleLoss}. Content Loss can be denoted as
\begin{equation}
\mathcal{L}_{c}^{j}=\frac{1}{C_{j}H_{j}W_{j}}\left \| \phi _{j}\left ( I_{out} \right)-\phi_{j}\left ( I_{gt} \right )  \right \|_{2}^{2}
\end{equation}
$\phi _{j}(x)$ is the j-th layer feature map of the pre-trained VGG-16 network. It has the shape of $(C_{j},H_{j},W_{j})$. Style Loss can be calculated as follow
\begin{equation}
\mathcal{L}_{s}^{j}=\left \| G_{j}^{\phi } (I_{out})- G_{j}^{\phi } (I_{gt}))\right \|_{F}^{2}
\end{equation}
It is the squared frobenius norm of two gram matrices $G_{j}^{\phi} (x)$. $G_{j}^{\phi} (x)$ is a matrix with the shape of $(C_{j},C_{j})$, and its elements are
\begin{equation}
G_{j}^{\phi}(x)_{c,c'}=\frac{1}{C_{j}H_{j}W_{j}}\sum_{h=1}^{H_{j}}\sum_{w=1}^{W_{j}}\phi _{j}(x)_{h,w,c}\phi _{j}(x)_{h,w,c'}
\end{equation}
Therefore, our enhanced loss can be recorded as
\begin{equation}
\mathcal{L}_{e}=\mathcal{L}_{c} + \lambda _{s}\mathcal{L}_{s}
\end{equation}
Where $\lambda _{s}$ are hyper-parameters, which will be discussed later.

\noindent\textbf{Multi-scale Loss.} Our correction is on feature-level. Ideally, the distortion of each layer features should be minimized. Therefore, we propose a multi-scale loss to further enhance the quality of the corrected feature maps. Particularly, multi-scale loss can be expressed by the following formula
\begin{equation}
\mathcal{L}_{m}=\sum_{i=1}^{L-1}\left \| S\left ( I_{gt},i \right ) - C\left ( I_{c}^{i}\oplus I_{d}^{i} \right ) \right \|_{1}
\end{equation}

$L$ is the number of convolution blocks. Specifically, we set $L$ to 6. $S$ is the downsampling operation. $S\left ( x,n \right )$ represents downsampling the input $x$ to the original $1/2^{n}$ times. $I_{d}^{i}$ and $I_{c}^{i}$ represent the original features at the decoder and the features corrected by feature correction layer, respectively. $\oplus $ denotes feature concatenation. $C$ is $3\times3$ convolution for decoding the output features into 3-channel RGB images. In this way, each feature map on the decoder can be effectively supervised.

\noindent\textbf{Overall Loss Function.} The loss function for training the complementary network is obtained by combining the reconstruction loss, adversarial loss, enhanced loss and multi-scale loss. We define the overall loss function as
\begin{equation}
\mathcal{L}=\lambda _{r}\mathcal{L}_{r}+\mathcal{L}_{adv}+\lambda _{m}\mathcal{L}_{m}+\mathcal{L}_{e}
\end{equation}
$\lambda _{r}$, $\lambda _{m}$ are hyper-parameters for balancing different loss functions. Through the overall loss function, we can achieve joint training of complementary networks.
\begin{figure}[!t]
\centering
\includegraphics[scale=.17]{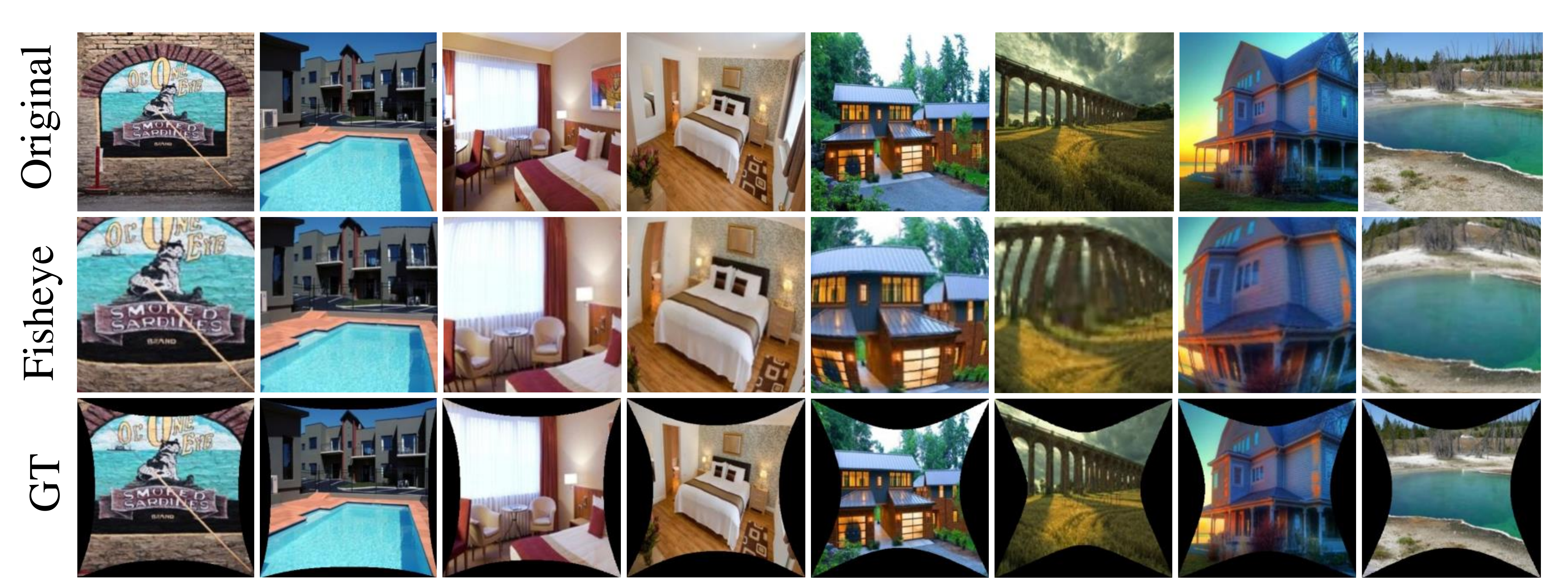}
\caption{
\label{dataset}
\textbf{Synthetic fisheye dataset.} Top: Original Place2 dataset \cite{Places2}. Middle: Generated fisheye images with different distortion. Bottom: Ground truth corresponding to the fisheye images.}
\vspace{-0.3cm}
\end{figure}

\begin{figure*}[!t]
\centering
\includegraphics[scale=.28]{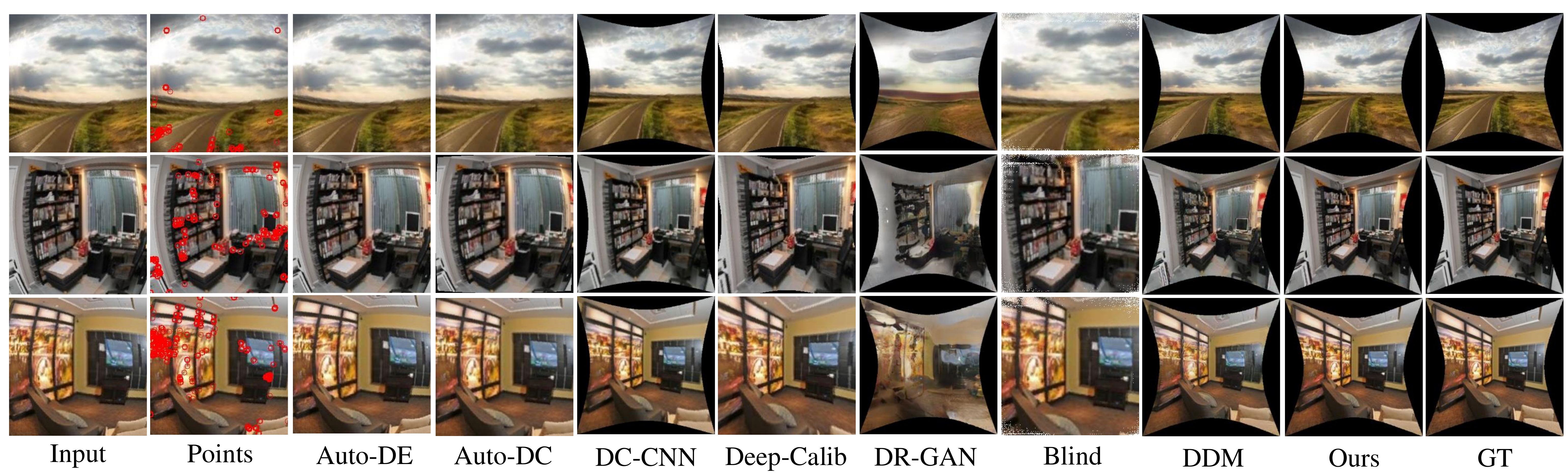}
\caption{
\label{result1}
\textbf{Comparison on synthetic images.} Visual results comparision in different scenarios. The state-of-the-art methods include: two traditional methods(Auto-DE \cite{Bukhari2013Automatic}, Auto-DC \cite{Aleman2014Automatic}), two regression-based methods(DC-CNN \cite{Rong2016Radial}, DeepCalib \cite{DeepCalib}), three generation-based methods(DR-GAN \cite{Liao2019}, Blind \cite{Blind}, DDM \cite{DDM}).}
\end{figure*}

\begin{table*}[!t]
\footnotesize
  \centering
  \caption{Comparison between the proposed method and the state-of-the-art methods on different image complexity.}
   \label{result table}
    \begin{tabular}{|l|c|c|c|c|c|c|c|c|c|c|c|c|}
    \hline
    \multicolumn{1}{|c|}{\multirow{1}[5]{*}{\diagbox[width=7.4em, height=2.2em, trim=l]{Methods}{Metrics}}} & \multicolumn{4}{c|}{$N \leq 200$ (30\%)} & \multicolumn{4}{c|}{$200 \leq N \leq 400$ (40\%)} & \multicolumn{4}{c|}{$N \geq 400$ (30\%) } \\
\cline{2-13}          & \multicolumn{1}{p{0.6cm}<{\centering}|}{PSNR} & \multicolumn{1}{p{0.6cm}<{\centering}|}{SSIM} & \multicolumn{1}{p{0.6cm}<{\centering}|}{FID} & \multicolumn{1}{p{1.2cm}<{\centering}|}{CW-SSIM} & \multicolumn{1}{p{0.6cm}<{\centering}|}{PSNR} & \multicolumn{1}{p{0.6cm}<{\centering}|}{SSIM} & \multicolumn{1}{p{0.6cm}<{\centering}|}{FID} & \multicolumn{1}{p{1.2cm}<{\centering}|}{CW-SSIM} & \multicolumn{1}{p{0.6cm}<{\centering}|}{PSNR} & \multicolumn{1}{p{0.6cm}<{\centering}|}{SSIM} & \multicolumn{1}{p{0.6cm}<{\centering}|}{FID} & \multicolumn{1}{p{1.2cm}<{\centering}|}{CW-SSIM}\\
    \hline
    \hline
    {Auto-DE \cite{Bukhari2013Automatic}} &   9.16    &  0.1964     &  301.9  & 0.4746  &  8.82     &  0.1478  & 328.7  &  0.4611     &  8.79     &   0.1073   & 366.5 & 0.4673  \\
    \hline
    {Auto-DC \cite{Aleman2014Automatic}} &   9.27    &  0.2005   &  298.5 &   0.4771    &  8.95      &  0.1538  & 325.9  &  0.4612     &  8.91    & 0.1129 &   361.6    & 0.4712 \\
    \Xhline{0.9pt}
    {DC-CNN \cite{Rong2016Radial}} &   13.83    &  0.4111     &  97.8  &  0.6375 &  13.42     &  0.3704     & 93.6   & 0.6249  &   13.01    &   0.3067   &97.0 & 0.6128 \\
    \hline
    {Deep-Calib \cite{DeepCalib}} &  20.59   &  0.6724     & 69.7  &  0.8383  & 18.34      &  0.5802  &  82.1  &   0.8063    &  18.83     &   0.5464   & 69.1 & 0.8129 \\
    \Xhline{0.9pt}
    {DR-GAN \cite{Liao2019}} &   17.23    &  0.5316     & 79.6  &  0.7794  &  16.57      &  0.5120   & 82.3 &  0.7528     &  16.24     & 0.5012  &  82.5   & 0.7387 \\
    \hline
    {Blind \cite{Blind}} &  20.00     & 0.7047   & 124.6  &   0.8343    &  19.01   & 0.6317 &  126.9     &  0.8153    & 18.62   &  0.6004 &  128.9     & 0.8139 \\
    \hline
    {DDM \cite{DDM}}   &  22.16     &   0.7538    & 55.4   & 0.9313  & 21.61      &  0.7339   & 59.3 &  0.9148     &  20.64    & 0.6875 &   59.9   & 0.9052 \\
    \Xhline{0.9pt}
    Ours  &  \textbf{25.06}     &  \textbf{0.8732}    & \textbf{25.1} &  \textbf{0.9615}   & \textbf{24.99}  & \textbf{0.8747}  &   \textbf{25.9}   &  \textbf{0.9648} &  \textbf{24.87} &  \textbf{0.8770}     &   \textbf{30.0}    & \textbf{0.9657}\\
    \hline
    \end{tabular}%
  \label{tab:addlabel}%
\vspace{-0.3cm}
\end{table*}%

\section{Experiments}
\subsection{Dataset and Implementation details}

\begin{figure}[!t]
\centering
\includegraphics[scale=.28]{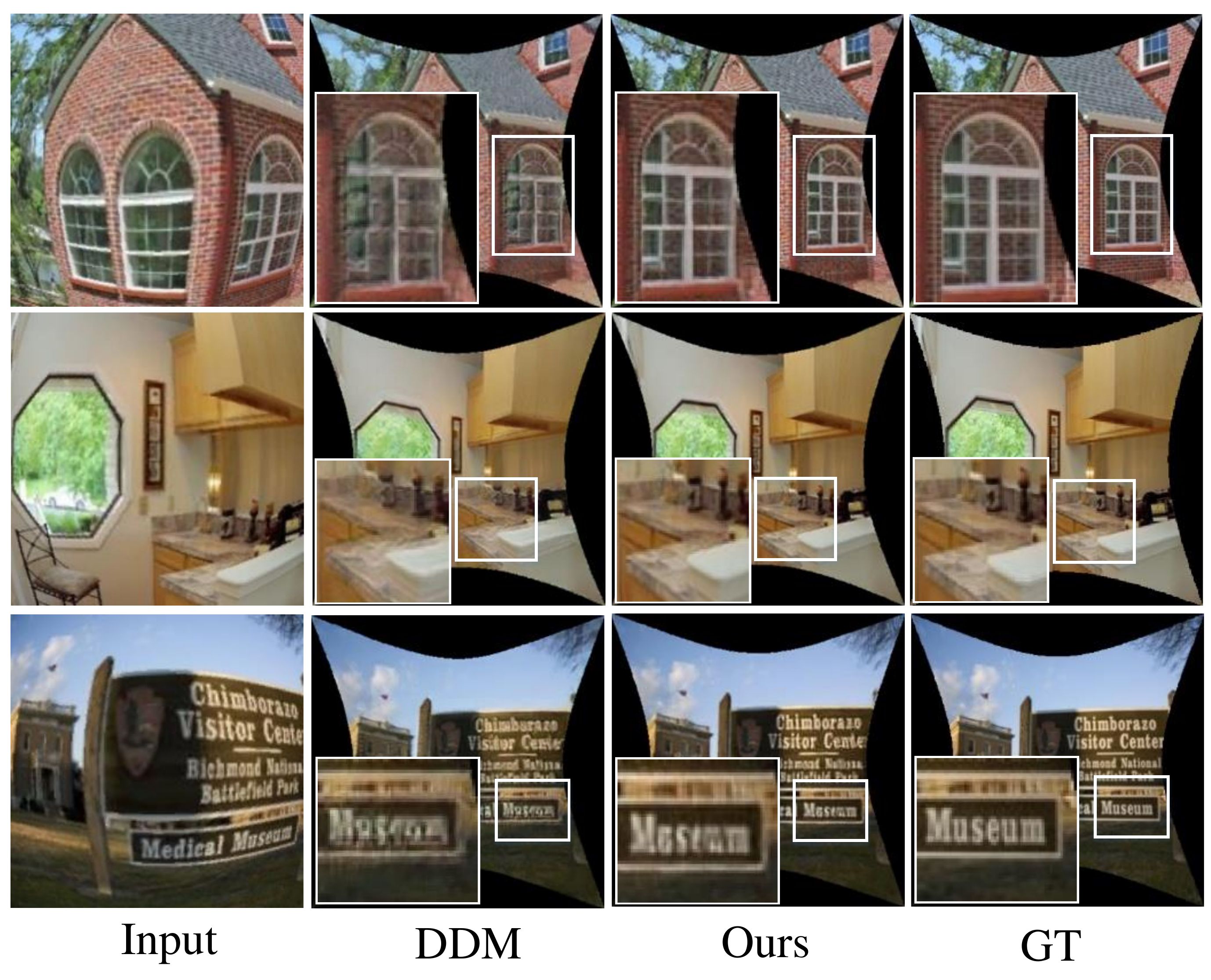}
\caption{
\label{zoom_result}
Additional comparison between DDM \cite{DDM} and our method. Our results provide more details.}
\vspace{-0.8cm}
\end{figure}

The fisheye dataset with corresponding labels is scarce. Therefore, we use a fisheye model to generate a synthetic fisheye dataset and select the Place2 dataset \cite{Places2} as our original data, as shown in Fig. \ref{dataset}. The Place2 dataset covers 400 types of scenes and contains 10 million images. We randomly select 100K pictures for training, 8k pictures for testing, and resize the generated fisheye image to $256\times256$. Like most existing rectification methods \cite{Yin2018FishEyeRecNet} \cite{Liao2019} \cite{Xue2019}, our polynomial model contains 4 parameters $P_{d}=\left [ k_{1}, k_{2}, k_{3}, k_{4} \right ]$. The value of $P_{d}$ selection is followed \cite{Liao2019} \cite{DDM} to obtain fisheye images with different distortion degree. We empirically set the hyper-parameters $\lambda _{r}$, $\lambda _{m}$, $\lambda _{s}$ to 60, 5, 2500 in overall loss function, respectively. In addition, we use the Adam algorithm with $\beta _{1}=0.5$ and $\beta _{2}=0.999$ to optimize our complementary network and set the initial learning rate to $10^{-4}$. NVIDIA GeForce GTX 2080Ti GPU is leveraged to train our network.

\subsection{Experimental Evaluation}
\label{Experimental results}

\begin{figure*}[!t]
\centering
\includegraphics[scale=.345]{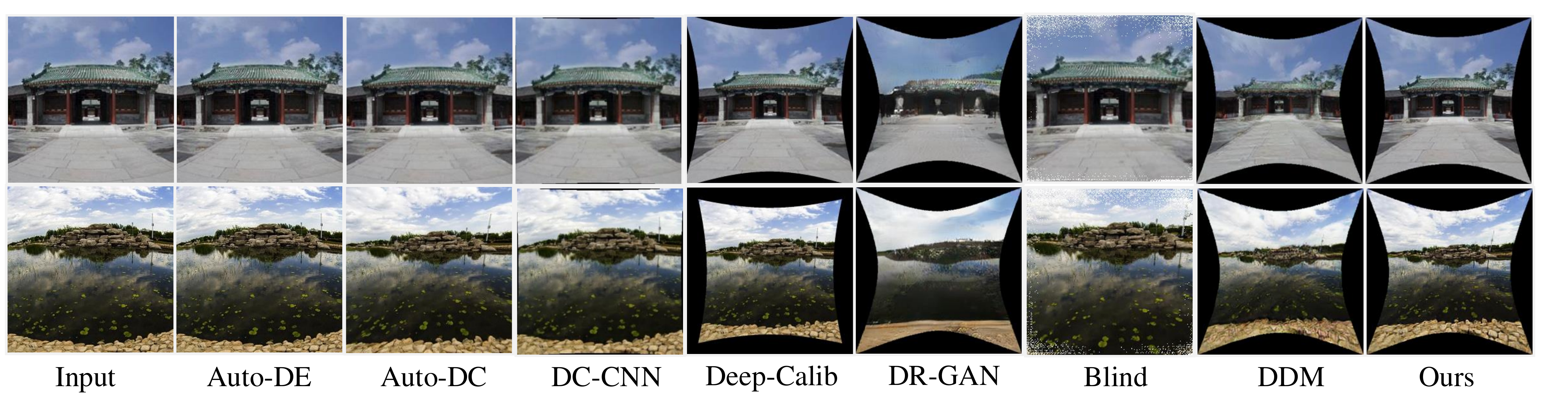}
\caption{
\label{real_result}
\textbf{Comparison on real fisheye images.} Our method outperforms the state-of-the-art methods both in corrected structure and image quality.}
\vspace{-0.5cm}
\end{figure*}

\noindent\textbf{Evaluation Matrics.} Peak Signal to Noise Ratio (PSNR) and Structural Similarity (SSIM) are the two most popular evaluation metrics for images. PSNR can effectively measure the detailed quality of the image, and SSIM can intuitively assess the image structure. In addition to PSNR and SSIM, Fr\'echet Inception Distance (FID) \cite{FID} can measure the difference between two distributions with the help of Wasserstein-2 distance, while complex wavelet structural similarity (CW-SSIM) \cite{CW-SSIM} can evaluate the quality on different geometric transformation. Therefore, we leverage them to objectively evaluate our experimental results.

\noindent\textbf{Comparing Methods.} To evaluate the performance, our method are compared with several state-of-the-art methods including: traditional methods(Auto-DE \cite{Bukhari2013Automatic}, Auto-DC \cite{Aleman2014Automatic}), regression-based methods(Deep-Calib \cite{DeepCalib}, DC-CNN \cite{Rong2016Radial}) and generation-based methods(Blind \cite{Blind}, DR-GAN \cite{Liao2019}, DDM \cite{DDM}).

\noindent\textbf{Quantitative Comparison Results.} Our quantitative comparison results are shown in Tab. \ref{result table}. To verify the robustness of the method, we leverage the harris algorithm to detect the interest points and divide the test dataset into 3 parts according to the number of corners detected. $N$ is the number of corners. The size of $N$ represents the complexity of the image scene. As reported in Tab. \ref{result table}, the traditional methods (Auto-DE \cite{Bukhari2013Automatic}, Auto-DC \cite{Aleman2014Automatic}) obtain poor performance. Regression-based methods(DC-CNN \cite{Rong2016Radial}, Deep-Calib \cite{DeepCalib}) are better than the traditional methods. Deep-Calib \cite{DeepCalib} also transcends the generation-based methods (Dr-GAN \cite{Liao2019}, Blind \cite{Blind}) except for DDM \cite{DDM}. Nevertheless, our method outperforms the above methods. In all the cases of $N \leq 200$, $200 \leq N \leq 400$, and $N \geq 400$, our method achieves the best performance, which fully proves the superiority of our method.

\noindent\textbf{Qualitative Comparison Results.} For a more intuitive comparison of corrected results, we visualize the results on the synthetic dataset (Fig. \ref{result1}). The correction effect of Auto-DE \cite{Bukhari2013Automatic} and Auto-DC \cite{Aleman2014Automatic} are not obvious for rarely detected features. Blind \cite{Blind} is difficult to correct images with large distortion. Deep-Calib \cite{DeepCalib} well corrects the center region of the image while degrades in the boundary regions. DC-CNN\cite{Rong2016Radial} does not fail in boundary regions, but the correction is not complete. DR-GAN\cite{Liao2019} is limited by the characteristics of the network itself, the generated image exhibits blur. DDM\cite{DDM} improves quality by superimposing different features to instruct the decoder. Although the structure correction of our method is similar to DDM\cite{DDM}, the corrected content of our method provides more details. As shown in Fig. \ref{zoom_result}, local regions of the generated images in DDM\cite{DDM} have poor quality, while our results provide richer texture information.

\begin{figure}[!t]
\centering
\includegraphics[scale=.50]{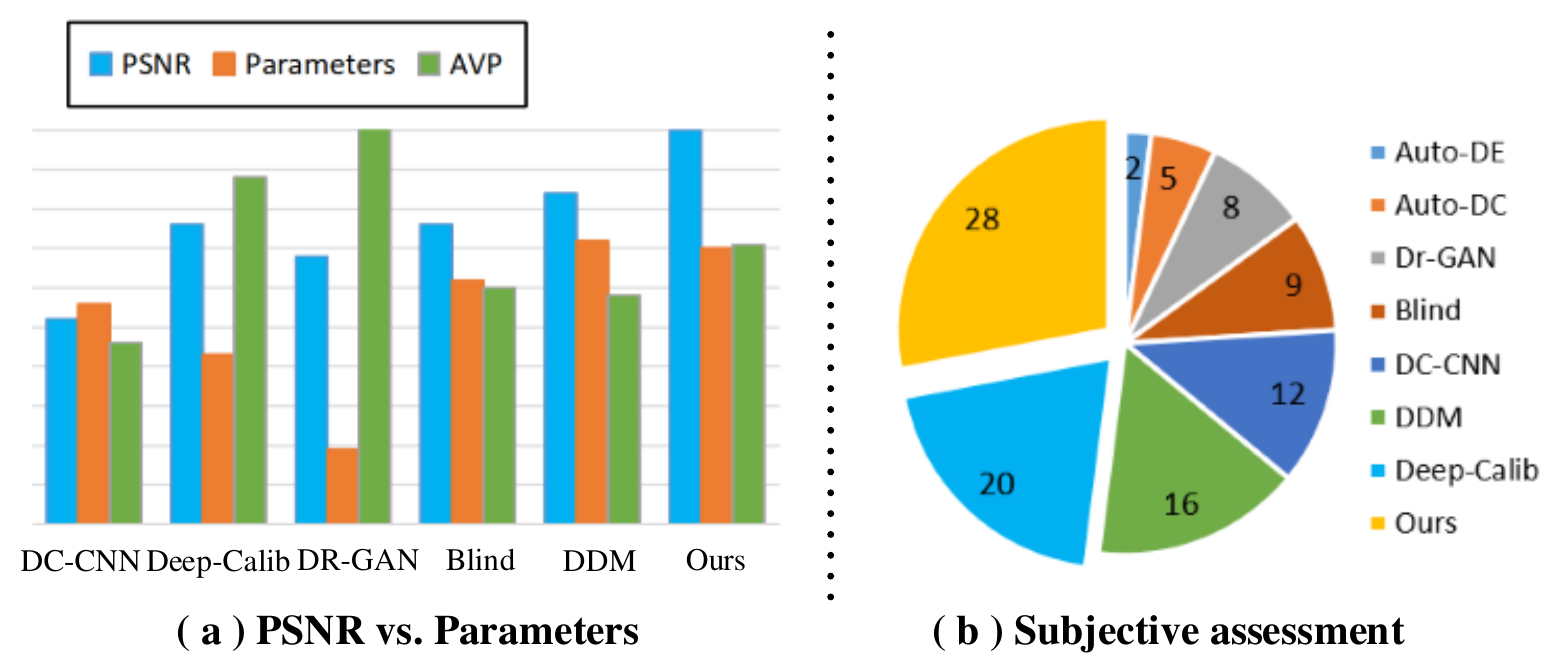}
\caption{
\label{ave}
\textbf{Network comparison and result evaluation.}  (a) A comparison of the PSNR, the number of parameters and AVP (average parameter performance) between deep learning correction methods. (b) Subjective assessment for the results.
}
\vspace{-0.4cm}
\end{figure}

\begin{figure}[!t]
\centering
\includegraphics[scale=.28]{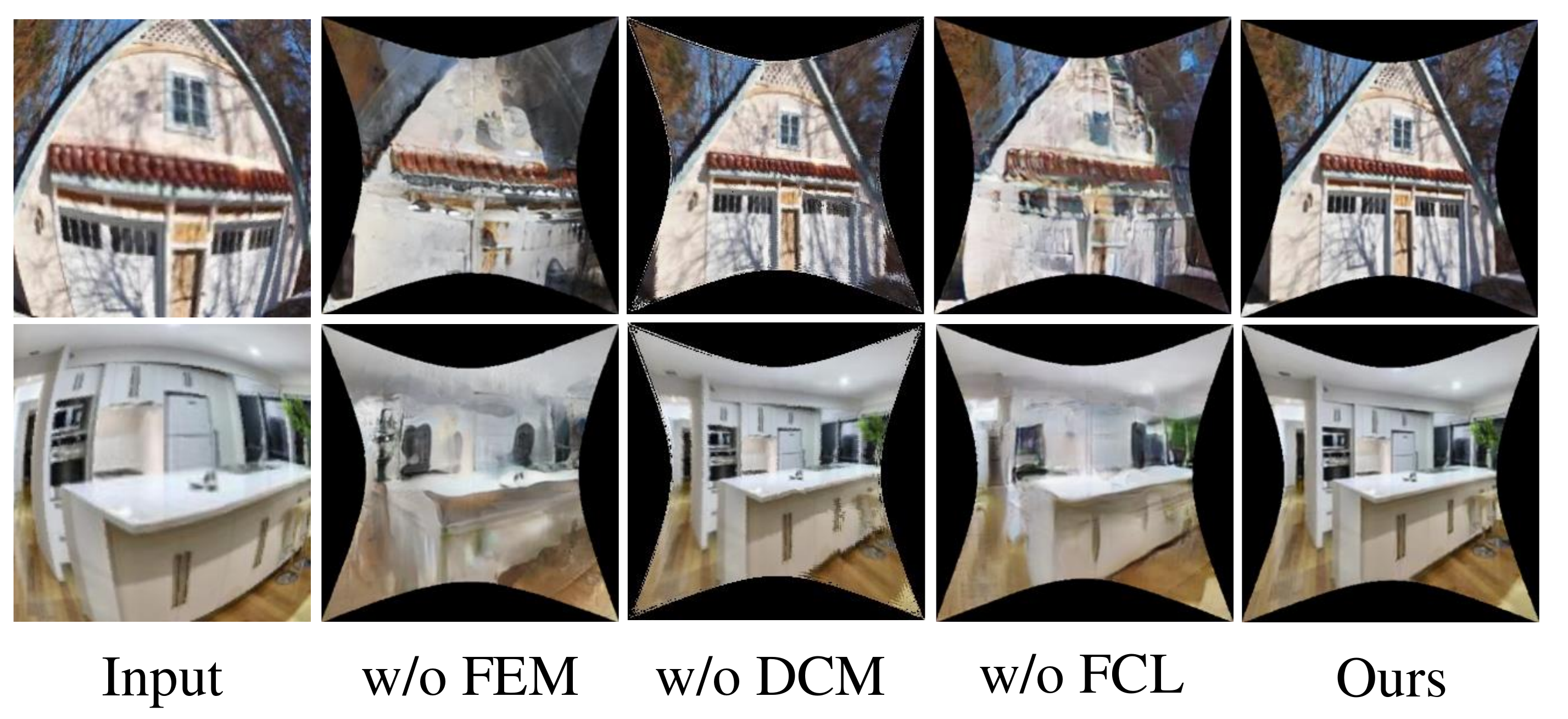}
\caption{
\label{ablation1}
A comparison of results based on different architecture.}
\vspace{-0.7cm}
\end{figure}

\begin{figure*}[!t]
\centering
\includegraphics[scale=.23]{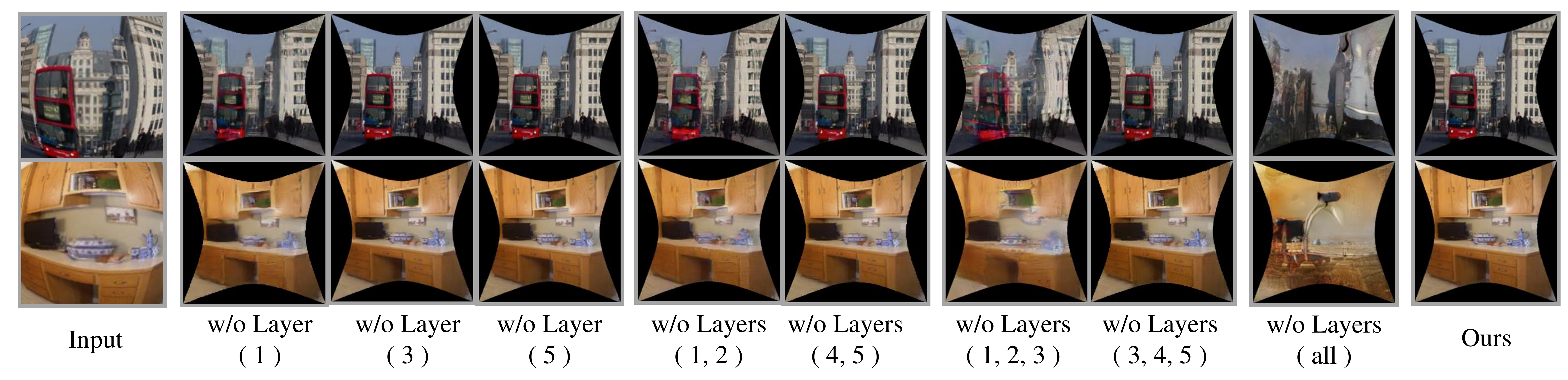}
\caption{
\label{ablation2}
A comparison results of without correcting image features on different layers.}
\vspace{-0.4cm}
\end{figure*}
\noindent\textbf{Comparison on real fisheye images.} Additional experimentation on real fisheye images is necessary. To further verify the practicality of the method, we leverage the synthetic data to train the network and then test the model on a real dataset. Compared with the state-of-the-art methods in Fig. \ref{real_result}, the corrected results of our method are subjectively better. Although there is a gap between the domain of synthetic dataset and real dataset, our method minimizes the gap by separating the content reconstruction from the structure correction and thereby achieves better performance.

\noindent\textbf{Average parameter performance.} We propose an average parameter performance (AVP), a novel evaluation metric, to reflect the performance improvement brought by each parameter. With the help of AVP, we can compare the efficiency of deep learning networks. AVP can be calculated by $PSNR/S$, where $S$ is the size of the network model. As shown in Fig. \ref{ave}\textcolor{red}{a}, our PSNR is in the best position and our AVP is in the third position, which proves that our network is effective and efficient.

\noindent\textbf{Subjective assessment.} We discover that some methods with lower quantitative performance have plausible visual results, like Deep-Calib \cite{DeepCalib}. Therefore, to further evaluate the results, we conduct a subjective assessment. We assigned 30 volunteers for the subjective assessment. They are required to select the best rectification results among all comparison methods from 100 fisheye image, which are randomly selected. Then, we averaged the votes of all volunteers and obtained the final subjective evaluation. As shown in Fig. \ref{ave}\textcolor{red}{b}, although Deep-Calib \cite{DeepCalib} outperforms other methods, the subjective assessment of our corrected results is still the best and exceeds Deep-Calib \cite{DeepCalib} method by 8\%.


\subsection{Ablation Study}
\label{Ablation Study}
To verify the effectiveness of each module, we decompose the network and then demonstrate its contribution.

\begin{table}[!t]
\footnotesize
\caption{Performance comparison for different structures and loss functions.}
  \label{structure ablation}
  \centering
    \begin{tabular}{p{1.8cm}<{}p{1.1cm}<{\centering}p{1.1cm}<{\centering}p{1.1cm}<{\centering}p{1.1cm}<{\centering}}
    \toprule
    \multicolumn{1}{c}{Methods} & PSNR  & SSIM  & \multicolumn{1}{c}{FID} & \multicolumn{1}{c}{CW-SSIM} \\
    \midrule
    w/o FEM & 16.09 & 0.4871 &   191.8    & 0.7431 \\
    w/o DCM & 19.12 & 0.6605 &   159.6  & 0.8784 \\
    w/o FCL & 21.31 & 0.7009 &   86.4  & 0.9187 \\
    \midrule
    w/o MS\&EH & 23.26 & 0.7872 &  36.2  & 0.9380 \\
    w/o EH & 23.69 & 0.7917 &   32.3  & 0.9400 \\
    \midrule
    Ours  & \textbf{24.98} & \textbf{0.8750} &  \textbf{26.9}     &  \textbf{0.9640}\\
    \bottomrule
    \end{tabular}%
  \label{tab:addlabel}%
\vspace{-0.4cm}
\end{table}%

\noindent\textbf{Structure And Loss Ablation.} The distortion correction module can independently correct the fisheye image. Therefore, we first remove the module, intending to verify the necessity of the flow estimation module (w/o FEM). Second, we remove the distortion correction module (w/o DCM) and feature correction layer (w/o FCL) respectively. The results are shown in Tab. \ref{structure ablation} and Fig. \ref{ablation1}. Simply superimposing the appearance flow improves the performance compared with the independent distortion correction module. However, with the help of the feature correction layer, the network can make full use of appearance flow to achieve more accurate correction. Besides, the results in Tab. \ref{structure ablation} prove that our loss functions multi-sale loss (MS) and enhanced loss (EH) can further improve network performance.

\noindent\textbf{Flow Ablation.} Our complementary network corrects the image features of all encoder layers, but the contribution of correcting layers is still ambiguous. Therefore, we analyze the impact as shown in Tab. \ref{flow ablation} and Fig. \ref{ablation2}. The results demonstrate the network has the worst performance without correcting all layers (w/o layers all). Correcting without the outermost layers (w/o layer 1) is worse than that correcting without the innermost layers (w/o layer 5). The reason is that compared to the inner image features (layer 5), the outer image features (layer 1) have greater distortion and more detailed information. Therefore, the outer image features (layer 1) need to be corrected more urgently than the inner image features (layer 5). In addition, we observe that the performance of correcting without the innermost layers (w/o layer 5) is similar to that of correcting all network layers (Ours). It further proves that the inner image features (layer 5) have slight distortion. Nonetheless, correcting them can also bring certain performance improvements.

\begin{table}[!t]
\footnotesize
  \centering
  \caption{Performance of using appearance flow on different convolutional layers.}
  \label{flow ablation}
    \begin{tabular}{p{2.0cm}<{}p{1.0cm}<{\centering}p{1.0cm}<{\centering}p{1.0cm}<{\centering}p{1.0cm}<{\centering}}
    \toprule
    \multicolumn{1}{c}{Methods} & \multicolumn{1}{c}{PSNR} & \multicolumn{1}{c}{SSIM} & \multicolumn{1}{c}{FID} & \multicolumn{1}{c}{CW-SSIM} \\
    \midrule
    w/o Layer(1) &  21.23     & 0.6878     &    90.3   & 0.9157 \\
    w/o Layer(3) &  22.95     & 0.7835      &   36.4    & 0.9365 \\
    w/o Layer(5) &  23.61     &  0.8064     &  33.9     &  0.9486\\
    \midrule
    w/o Layers(1,2) &  21.20     &  0.6831     &  87.5  & 0.9164 \\
    w/o Layers(4,5) &  23.16     &  0.8031     &   33.6 &  0.9426\\
    \midrule
    w/o Layers(1,2,3) &  19.64     &  0.6048     &  132.7     & 0.8823 \\
    w/o Layers(3,4,5) &  22.63     &  0.7812     &  40.1     & 0.9362 \\
    \midrule
    w/o Layers(all) & 16.09      & 0.4871      &   191.8    & 0.7431 \\
    \midrule
    Ours  &  \textbf{24.98}     &  \textbf{0.8750}     &  \textbf{26.9}     & \textbf{0.9640}  \\
    \bottomrule
    \end{tabular}%
   \vspace{-0.7cm}
\end{table}%

\section{Conclusion}
In this paper, we propose a progressively complementary network to correct fisheye images. Two modules are connected in parallel, which can correct the fisheye image and estimate the distortion structure simultaneously. Different from the existing generation-based methods, we uniquely insert the feature correction layer into the skip connection in our network. Pre-correction is implemented before transferring the features, which fundamentally solves the problem of distortion diffusion and implements a feature-level correction. Particularly, taking advantage of the progressive generation characteristics, we design two modules as a novel complementary structure and introduce a multi-scale loss function to supervise the corrected image features. It further enhances the quality of the corrected image. The experimental results of our network significantly outperform the state-of-the-art methods, both subjectively and objectively.

\noindent\textbf{Acknowledgments:} This work was supported in part by Fundamental Research Funds for the Central Universities (2020YJS028), in part by National Natural Science Foundation of China (No.61772066, No.62072026) and Beijing Natural Science Foundation (JQ20022).

{\small
\bibliographystyle{unsrt}
\bibliographystyle{ieee_fullname}

\begin{thebibliography}{10}

\bibitem{Muhammad2019EfficientDC}
M.~Khan, J.~Aḥmad, Z.~Lv, P.~Bellavista, P.~Yang, and S.~Baik.
\newblock Efficient deep cnn-based fire detection and localization in video
  surveillance applications.
\newblock {\em IEEE Transactions on Systems, Man, and Cybernetics: Systems},
  49:1419--1434, 2019.

\bibitem{Geiger2012AreWR}
G.~Andreas, L.~Philip, and R.~Urtasun.
\newblock Are we ready for autonomous driving? the kitti vision benchmark
  suite.
\newblock {\em CVPR}, pages 3354--3361, 2012.

\bibitem{Mahmoodi2019OptimalJS}
S.~E. Mahmoodi, R.~N. Uma, and K.~P. Subbalakshmi.
\newblock Optimal joint scheduling and cloud offloading for mobile
  applications.
\newblock {\em IEEE Transactions on Cloud Computing}, 7:301--313, 2019.

\bibitem{tracking}
S.~Zhang, H.~Yao, X.~Sun, and X.~Lu.
\newblock Sparse coding based visual tracking: Review and experimental
  comparison.
\newblock {\em Pattern Recognition}, 46:1772--1788, 2013.

\bibitem{Li2019TargetAwareDT}
X.~Li, C.~Ma, B.~Wu, Z.~He, and M.~Yang.
\newblock Target-aware deep tracking.
\newblock {\em CVPR}, pages 1369--1378, 2019.

\bibitem{motionestimate}
T.~J. {Broida}, S.~Chandrashekhar, and R.~Chellappa.
\newblock Recursive 3-d motion estimation from a monocular image sequence.
\newblock {\em IEEE Transactions on Aerospace and Electronic Systems},
  26(4):639--656, 1990.

\bibitem{Bao2019MEMCNetME}
W.~Bao, W.~Lai, X.~Zhang, Z.~Gao, and M.~Yang.
\newblock Memc-net: Motion estimation and motion compensation driven neural
  network for video interpolation and enhancement.
\newblock {\em IEEE transactions on pattern analysis and machine intelligence},
  2019.

\bibitem{Scene_Seg}
J.~Manuel {\'A}lvarez, T.~Gevers, Y.~LeCun, and A.~M. L{\'o}pez.
\newblock Road scene segmentation from a single image.
\newblock In {\em ECCV}, 2012.

\bibitem{Fu2019DualAN}
J.~Fu, J.~Liu, T.~Haijie, Z.~Fang, and H.~Lu.
\newblock Dual attention network for scene segmentation.
\newblock {\em CVPR}, pages 3141--3149, 2019.

\bibitem{Rui2014Unsupervised}
R.~Melo, M.~Antunes, J.~Pedro Barreto, G.~Falc{\~a}o~Paiva Fernandes, and
  N.~Gonçalves.
\newblock Unsupervised intrinsic calibration from a single frame using a
  "plumb-line" approach.
\newblock {\em ICCV}, pages 537--544, 2013.

\bibitem{Bukhari2013Automatic}
F.~Bukhari and M.~N. Dailey.
\newblock Automatic radial distortion estimation from a single image.
\newblock {\em Journal of Mathematical Imaging \& Vision}, 45(1):31--45, 2013.

\bibitem{Zhang2015}
M.~Zhang, J.~Yao, M.~Xia, K.~Li, Y.~Zhang, and Y.~Liu.
\newblock Line-based multi-label energy optimization for fisheye image
  rectification and calibration.
\newblock {\em CVPR}, pages 4137--4145, 2015.

\bibitem{Barreto2005}
J.~Pedro Barreto and H.~Sabino de~Ara{\'u}jo.
\newblock Geometric properties of central catadioptric line images and their
  application in calibration.
\newblock {\em IEEE Transactions on Pattern Analysis and Machine Intelligence},
  27:1327--1333, 2005.

\bibitem{Rong2016Radial}
J.~Rong, S.~Huang, Z.~Shang, and X.~Ying.
\newblock Radial lens distortion correction using convolutional neural networks
  trained with synthesized images.
\newblock In {\em ACCV}, 2016.

\bibitem{DeepCalib}
O.~Bogdan, V.~Eckstein, F.~Rameau, and J.~Bazin.
\newblock Deepcalib: a deep learning approach for automatic intrinsic
  calibration of wide field-of-view cameras.
\newblock In {\em CVMP}, 2018.

\bibitem{Yin2018FishEyeRecNet}
X.Yin, X.~Wang, J.~Yu, M.~Zhang, P.~Fua, and D.~Tao.
\newblock Fisheyerecnet: A multi-context collaborative deep network for fisheye
  image rectification.
\newblock In {\em ECCV}, pages 475--490, 2018.

\bibitem{Xue2019}
Z.~Xue, N., G.~Xia, and W.~Shen.
\newblock Learning to calibrate straight lines for fisheye image rectification.
\newblock {\em CVPR}, pages 1643--1651, 2019.

\bibitem{Liao2019}
K.~Liao, C.~Lin, Y.~Zhao, and M.~Gabbouj.
\newblock {DR-GAN}: Automatic radial distortion rectification using conditional
  {GAN} in real-time.
\newblock {\em IEEE Transactions on Circuits and Systems for Video Technology},
  2019.

\bibitem{DDM}
K.~Liao, C.~Lin, Y.~Zhao, and M.~Xu.
\newblock Model-free distortion rectification framework bridged by distortion
  distribution map.
\newblock {\em IEEE Transactions on Image Processing}, 29:3707--3718, 2020.

\bibitem{Blind}
X.~Li, B.~Zhang, Pedro~V. Sander, and J.~Liao.
\newblock Blind geometric distortion correction on images through deep
  learning.
\newblock In {\em CVPR}, pages 4855--4864, 2019.

\bibitem{STM}
M.~Jaderberg, K.~Simonyan, A.~Zisserman, and K.~Kavukcuoglu.
\newblock Spatial transformer networks.
\newblock In {\em NIPS}, pages 2017--2025, 2015.

\bibitem{ZHANG1999}
Z.~Zhang.
\newblock Flexible camera calibration by viewing a plane from unknown
  orientations.
\newblock {\em ICCV}, 1:666--673 vol.1, 1999.

\bibitem{Mei2007}
C.~Mei and P.~Rives.
\newblock Single view point omnidirectional camera calibration from planar
  grids.
\newblock {\em IEEE International Conference on Robotics and Automation}, pages
  3945--3950, 2007.

\bibitem{Gasparini2009}
S.~Gasparini, P.~F. Sturm, and J.~Pedro Barreto.
\newblock Plane-based calibration of central catadioptric cameras.
\newblock {\em ICCV}, pages 1195--1202, 2009.

\bibitem{Puig2010}
L.~Puig, Y.~Bastanlar, P.~Sturm, J.~J. Guerrero, and J.~Barreto.
\newblock Calibration of central catadioptric cameras using a dlt-like
  approach.
\newblock {\em International Journal of Computer Vision}, 93:101--114, 2010.

\bibitem{Zhang2000AFN}
Z.~Zhang.
\newblock A flexible new technique for camera calibration.
\newblock {\em IEEE Trans. Pattern Anal. Mach. Intell.}, 22:1330--1334, 2000.

\bibitem{Dansereau2013DecodingCA}
D.~Dansereau, O.~Pizarro, and S.~B. Williams.
\newblock Decoding, calibration and rectification for lenselet-based plenoptic
  cameras.
\newblock {\em 2013 IEEE Conference on Computer Vision and Pattern
  Recognition}, pages 1027--1034, 2013.

\bibitem{Stuiver1993Extended1D}
M.~Stuiver and P.~Reimer.
\newblock Extended 14c data base and revised calib 3.0 14c age calibration
  program.
\newblock {\em Radiocarbon}, 35:215--230, 1993.

\bibitem{Chander2009SummaryOC}
G.~Chander, B.~Markham, and D.~Helder.
\newblock Summary of current radiometric calibration coefficients for landsat
  mss, tm, etm+, and eo-1 ali sensors.
\newblock {\em Remote Sensing of Environment}, 113:893--903, 2009.

\bibitem{Geiger2012AutomaticCA}
G.~Andreas, F.~Moosmann, C.~Omer, and B.~Schuster.
\newblock Automatic camera and range sensor calibration using a single shot.
\newblock {\em 2012 IEEE International Conference on Robotics and Automation},
  pages 3936--3943, 2012.

\bibitem{StraightLine}
F.~Devernay and O.~Faugeras.
\newblock Straight lines have to be straight: automatic calibration and removal
  of distortion from scenes of structured enviroments.
\newblock {\em Machine Vision Applications}, 13(1):14--24, 2001.

\bibitem{Guo2017OnCO}
C.~Guo, P.~Geoff, Y.~Sun, and K.~Q. Weinberger.
\newblock On calibration of modern neural networks.
\newblock {\em ArXiv}, abs/1706.04599, 2017.

\bibitem{Zhang2015LargeScaleAS}
W.~Zhang, H.~Ren, C.~Pan, M.~Chen, R.~C.~D. Lamare, B.~Du, and J.~Dai.
\newblock Large-scale antenna systems with ul/dl hardware mismatch: Achievable
  rates analysis and calibration.
\newblock {\em IEEE Transactions on Communications}, 63:1216--1229, 2015.

\bibitem{Lee2019CorrectionOB}
M.~Lee, K.~Hyungtae, and P.~Joonki.
\newblock Correction of barrel distortion in fisheye lens images using
  image-based estimation of distortion parameters.
\newblock {\em IEEE Access}, 7:45723--45733, 2019.

\bibitem{Yang2010CorrectAP}
M.~Yang, C.~Yang, and T.~Meng.
\newblock Correct a particular fisheye lens distortion quickly using the
  coordinate map table.
\newblock In {\em International Conference on Image Processing and Pattern
  Recognition in Industrial Engineering}, 2010.

\bibitem{Grant2015AutomaticEA}
G.~Timothy and N.~Grigorieff.
\newblock Automatic estimation and correction of anisotropic magnification
  distortion in electron microscopes.
\newblock {\em Journal of structural biology}, 192 2:204--8, 2015.

\bibitem{Ophus2016CorrectingND}
C.~Ophus, J.~Ciston, and C.~T. Nelson.
\newblock Correcting nonlinear drift distortion of scanning probe and scanning
  transmission electron microscopies from image pairs with orthogonal scan
  directions.
\newblock {\em Ultramicroscopy}, 162:1--9, 2016.

\bibitem{Krizhevsky2017ImageNetCW}
A.~Krizhevsky, S.~Ilya, and G.~E. Hinton.
\newblock Imagenet classification with deep convolutional neural networks.
\newblock In {\em CACM}, 2017.

\bibitem{Ledig2017PhotoRealisticSI}
C.~Ledig, L.~Theis, F.~Husz{\'a}r, J.~Caballero, A.~Andrew, T.~Alykhan,
  J.~Totz, Z.~Wang, and W.~Shi.
\newblock Photo-realistic single image super-resolution using a generative
  adversarial network.
\newblock {\em CVPR}, pages 105--114, 2017.

\bibitem{Fitzgibbon2001}
A.~W. Fitzgibbon.
\newblock Simultaneous linear estimation of multiple view geometry and lens
  distortion.
\newblock {\em CVPR}, 1:I--I, 2001.

\bibitem{Basu1995}
A.~Basu and S.~Licardie.
\newblock Alternative models for fish-eye lenses.
\newblock {\em Pattern Recognition Letters}, 16:433--441, 1995.

\bibitem{StyleLoss}
J.~Johnson, A.~Alahi, and F.~Li.
\newblock Perceptual losses for real-time style transfer and super-resolution.
\newblock In {\em ECCV}, pages 694--711, 2016.

\bibitem{Places2}
B.~Zhou, A.~Lapedriza, A.~Khosla, A.~Oliva, and A.~Torralba.
\newblock Places: A 10 million image database for scene recognition.
\newblock {\em IEEE Transactions on Pattern Analysis and Machine Intelligence},
  40:1452--1464, 2018.

\bibitem{Aleman2014Automatic}
M.~Alem{\'a}nflores, L.~{\'A}lvarez, L.~G{\'o}mez, and D.SantanaCedr{\'e}s.
\newblock Automatic lens distortion correction using one-parameter division
  models.
\newblock {\em IPOL}, 4:327--343, 2014.

\bibitem{FID}
H.~Martin, R.~Hubert, U.~Thomas, N.~Bernhard, and S.~Hochreiter.
\newblock Gans trained by a two time-scale update rule converge to a local nash
  equilibrium.
\newblock In {\em NIPS}, 2017.

\bibitem{CW-SSIM}
M.~P. Sampat, Z.~Wang, S.~Gupta, A.~C. Bovik, and M.~K. Markey.
\newblock Complex wavelet structural similarity: A new image similarity index.
\newblock {\em IEEE Transactions on Image Processing}, 18(11):2385--2401, 2009.

\end{thebibliography}
}

\end{document}